%% file: root.tex
\documentclass[a4paper,conference]{IEEEtran}
\usepackage{cite}
\usepackage{amsmath,amssymb,amsfonts}
\usepackage{algorithmic}
\usepackage{graphicx}
\usepackage{textcomp}
\usepackage{xcolor}
\usepackage{multirow}
\def\BibTeX{{\rm B\kern-.05em{\sc i\kern-.025em b}\kern-.08em
    T\kern-.1667em\lower.7ex\hbox{E}\kern-.125emX}}
\begin{document}
\IEEEoverridecommandlockouts
\title{Mutual-Supervised Feature Modulation Network for Occluded Pedestrian Detection}

\author{\IEEEauthorblockN{Ye He, Chao Zhu$^*$, Xu-Cheng Yin}
\IEEEauthorblockA{School of Computer and Communication Engineering,\\
University of Science and Technology Beijing \\
Beijing, China \\
Email: s20190676@xs.ustb.edu.cn, chaozhu@ustb.edu.cn, xuchengyin@ustb.edu.cn}
\thanks{*Corresponding author}}
\maketitle

\input{abstract}
\input{introduction}
\input{related}
\input{method}

\input{experiment}
\input{conclusion}

\input{acknowledgment}

\IEEEpeerreviewmaketitle

{\small
	\bibliographystyle{unsrt}
	\bibliography{ref}
}

\end{document}

%% file: abstract.tex
\begin{abstract}

State-of-the-art pedestrian detectors have achieved significant progress on non-occluded pedestrians, yet they are still struggling under heavy occlusions. The recent occlusion handling strategy of popular two-stage approaches is to build a two-branch architecture with the help of additional visible body annotations. Nonetheless, these methods still have some weaknesses. Either the two branches are trained independently with only score-level fusion, which cannot guarantee the detectors to learn robust enough pedestrian features. Or the attention mechanisms are exploited to only emphasize on the visible body features. However, the visible body features of heavily occluded pedestrians are concentrated on a relatively small area, which will easily cause missing detections. To address the above issues, we propose in this paper a novel Mutual-Supervised Feature Modulation (MSFM) network, to better handle occluded pedestrian detection. The key MSFM module in our network calculates the similarity loss of full body boxes and visible body boxes corresponding to the same pedestrian so that the full-body detector could learn more complete and robust pedestrian features with the assist of contextual features from the occluding parts. To facilitate the MSFM module, we also propose a novel two-branch architecture, consisting of a standard full body detection branch and an extra visible body classification branch. These two branches are trained in a mutual-supervised way with full body annotations and visible body annotations, respectively. To verify the effectiveness of our proposed method, extensive experiments are conducted on two challenging pedestrian datasets: Caltech and CityPersons, and our approach achieves superior performance compared to other state-of-the-art methods on both datasets, especially in heavy occlusion cases.

\end{abstract}

%% file: introduction.tex
\section{Introduction}
Pedestrian detection is a challenging computer vision task that has been widely applied in numerous applications, such as autonomous driving, robotics, and intelligent video surveillance. State-of-the-art pedestrian detectors~\cite{ren2017accurate,liu2018learning,cai2016unified,mao2017can,brazil2017illuminating,wang2018repulsion} have achieved significant progress on non-occluded pedestrians, yet they are still confronted by heavy occlusions. For example, when walking on a street, pedestrians are likely to be occluded by other pedestrians or other objects, such as cars, buildings, and bicycles. Therefore, it remains as one of the most challenging issues for a pedestrian detection approach to robustly detect partially or heavily occluded pedestrians. 

\begin{figure}[htb]
\centering
\includegraphics[width=\linewidth]{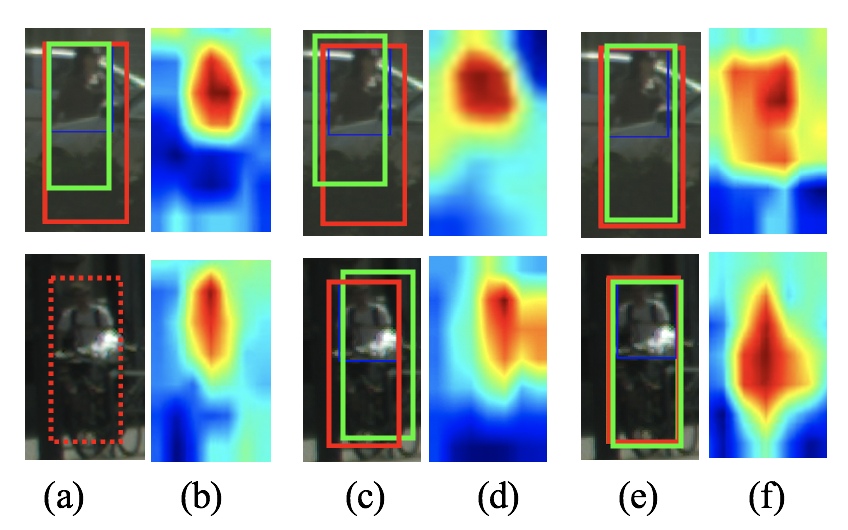}
\caption{Visual comparison between Bi-box, MGAN and MSFMN. (a), (c), and (e) represent the detection results of different methods. (b), (d), and (f) represent the feature visualization. Solid red boxes represent full body annotations, blue boxes are visible body annotations, green boxes denote detection results, and dashed red boxes represent the missed detections. The detected regions are cropped from the corresponding images in CityPersons val. set. Compared with Bi-box and MGAN, MSFMN displays a high response not only on the visible part but also on the occluding part.}
\label{fig:featureVisualization}
\end{figure}

Many existing approaches~\cite{ren2017accurate,liu2018learning,cai2016unified,mao2017can} employ a simple detection strategy that assumes entirely visible pedestrians when trained with full body annotations. Despite achieving impressive results for non-occluded pedestrians, such a strategy is still struggling under partial or heavy occlusions since the features of the occluding part are vastly different from the visible part. 

Compared with full body regions, visible parts of pedestrians usually suffer much less from occlusion, which can provide more discriminative and confident cues. Several recent approaches~\cite{zhou2018bi,zhang2018occluded,pang2019mask} deal with occlusions by building a two-branch architecture with extra visible-region information, available with standard pedestrian detection benchmarks, like Caltech~\cite{dollar2011pedestrian} and CityPersons~\cite{zhang2017citypersons}.
Nonetheless, these methods still have some weaknesses. Either the two branches are trained independently with only score-level fusion, which cannot guarantee the detectors to learn robust enough pedestrian features, such as Bi-box~\cite{zhou2018bi}. Or the attention mechanisms are exploited to emphasize on the visible regions while suppressing the occluded regions, like MGAN~\cite{pang2019mask}. For these methods, the visible body features of the heavily occluded pedestrians are concentrated on a relatively small area, which will easily cause missing detections. Besides, some two-branch methods collect positive training samples with full body annotations and visible body annotations simultaneously, which may sacrifice some useful visible features, such as Bi-box~\cite{zhou2018bi}. As illustrated in Fig. \ref{fig:featureVisualization}, (b) and (d) depict the feature maps learned by Bi-box~\cite{zhou2018bi} and MGAN~\cite{pang2019mask}, respectively. It can be observed that only the visible part has a high response while the occluded part almost has no response, and this will easily lead to inaccurate, even missing detections in heavy occlusion cases. Therefore, we think it is insufficient to only focus on the features within the visible bounding boxes in the case of heavy occlusions. The assist of the features from the occluding part is also important as a contextual cue to enhance pedestrian detectors against heavy occlusions, and this has not been studied thoroughly in previous works.

\begin{figure}[!t]
\centering
\includegraphics[width=\linewidth]{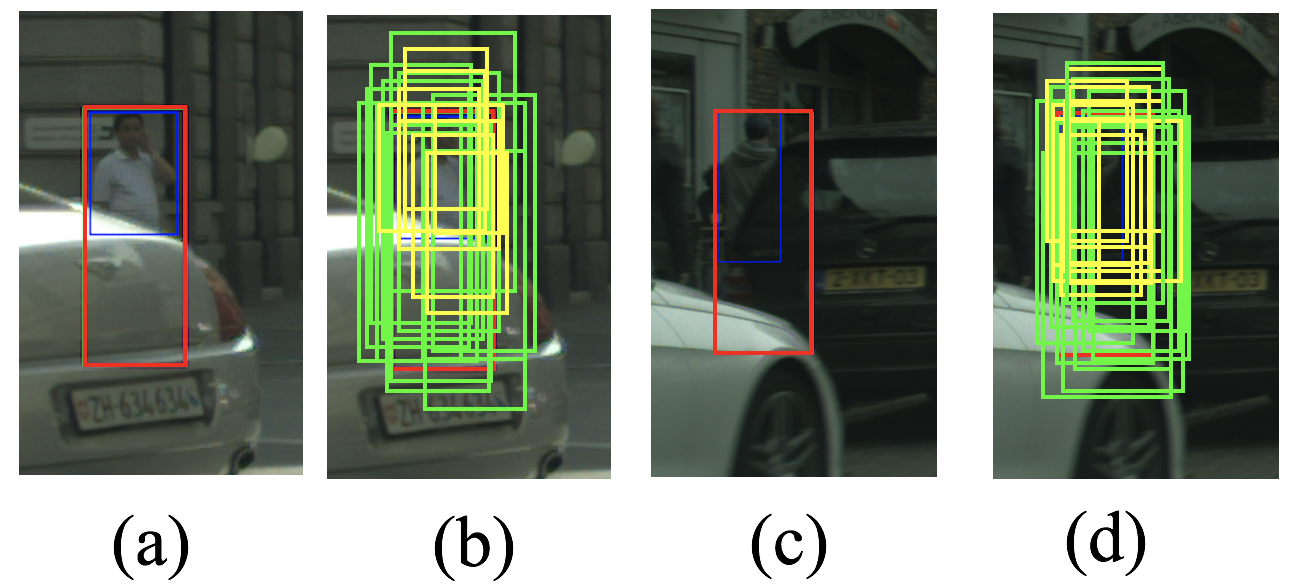}
\caption{Visualization of training samples obtained by our proposed sampling method. (a) and (c) are training images in CityPersons train. set. The Red boxes denote full body ground truth box and the blue boxes represent the visible body ground truth box of a pedestrian. (b) and (d) depict the positive training samples. Green boxes are positive training samples collected from full body branch and yellow boxes denote positive training samples collected from visible body branch.}
\label{fig:samples}
\end{figure}

To this end, we propose in this paper a novel Mutual-Supervised Feature Modulation Network (MSFMN) aiming at enhancing feature representations of occluded pedestrians. A key part of the proposed MSFMN is the Mutual-Supervised Feature Modulation module, which aims to calculate the similarity loss between full body boxes and visible body boxes corresponding to the same pedestrian. Therefore, the full-body detector could learn more complete and robust pedestrian features in a mutual-supervised way with the assist of contextual features from the occluding parts. To obtain the visible boxes, we also construct a novel two-branch architecture consisting of a standard full body detection branch and an extra visible body classification branch. Moreover, these two branches sample their training samples supervised by full body annotations and visible body annotations, respectively (as displayed in Eq.1 and Eq.2) to obtain more focused training samples as shown in Fig. \ref{fig:samples}. Note that the proposed method can be easily applied to any existing two-stage detection framework. Finally, as shown in Fig. \ref{fig:featureVisualization} (f), it is obvious that the feature responses of the proposed method are concentrated on a relatively wider region, including not only the visible part but also the occluding part. Therefore, the contextual information from the occluding regions can essentially enhance the discriminability of the features of occluded pedestrians.

In summary, the main contributions of this paper are three-fold:

(1)	We propose a novel Mutual-Supervised Feature Modulation Network (MSFMN), to deal with the problem of occluded pedestrian detection. The key Mutual-Supervised Feature Modulation module calculates the similarity loss between full body boxes and visible body boxes to learn more robust feature representations of occluded pedestrians;

(2)	The MSFMN comprises two branches: a standard full body detection branch and an extra visible body classification branch. These two branches are supervised by full body annotations and visible body annotations, respectively, to obtain more focused training samples;

(3)	Extensive experiments are conducted on two standard pedestrian detection benchmarks: CityPersons~\cite{zhang2017citypersons} and Caltech~\cite{dollar2011pedestrian}. Our approach sets a new state-of-the-art on both datasets, strongly validating the effectiveness of the proposed method for occluded pedestrian detection.

\begin{figure*}[t!]
\centering
\includegraphics[width=\linewidth]{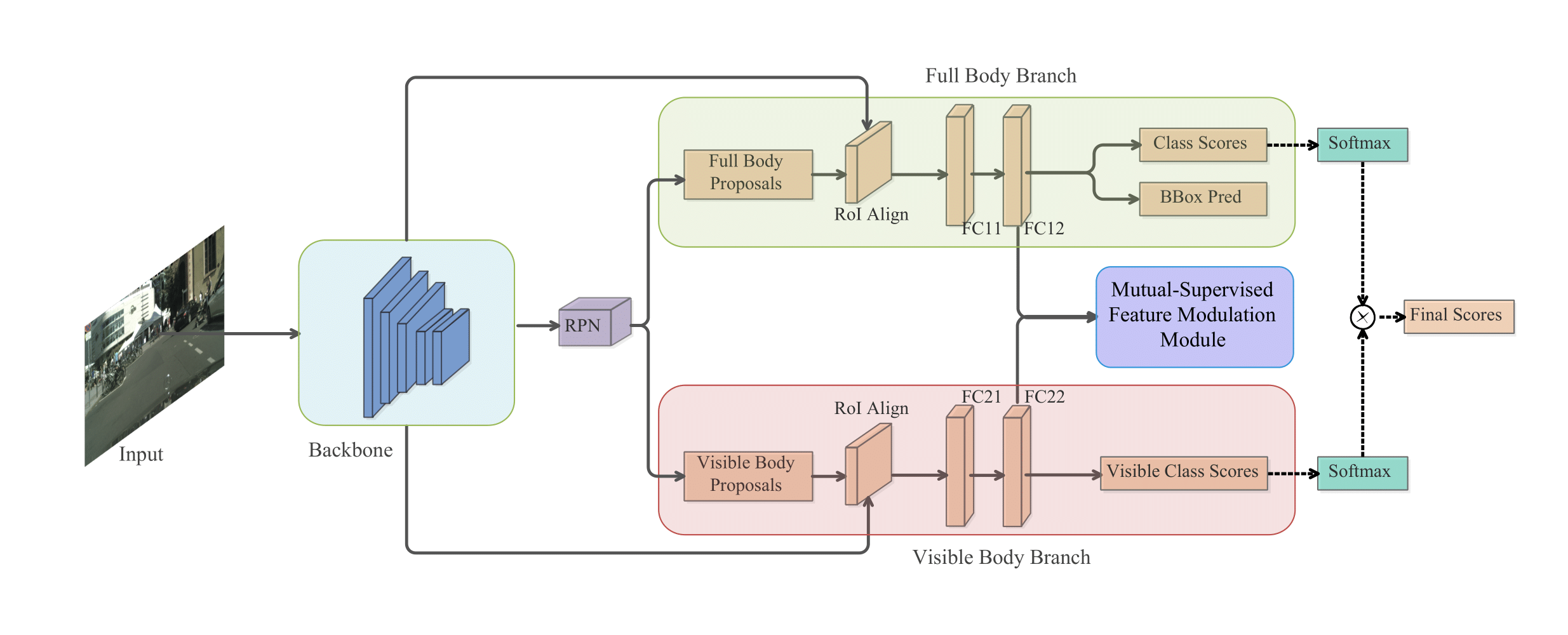}

\caption{Overall network architecture of our Mutual-Supervised Feature Modulation Network (MSFMN). It consists of a full body (FB) branch enclosed in the green box and a visible body (VB) branch in the red box. A novel Mutual-Supervised Feature Modulation Module is enclosed in the purple box. Two feature vectors are obtained from fully connected layer FC12 and FC22 respectively and then sent to the Mutual-Supervised Feature Modulation Module. In the architecture, the FB branch is a standard pedestrian detector branch and the VB branch is proposed to generate classification scores for visible proposals. FC$_{ij}$ denotes the j-${th}$ FC layer in the i-${th}$ branch. The dotted lines depict the inference process. $\otimes$ represents element-wise product operation.}
%
\label{fig:overall}
\end{figure*}

%% file: related.tex
\section{Related Work}

\subsection{Deep Pedestrian Detection}

With the rapid development of convolutional neural networks (CNNs)~\cite{simonyan2014very, he2016deep,hu2018squeeze}, great progress has been made in the pedestrian detection field. Most existing CNN-based pedestrian detectors employ either one-stage or two-stage strategy as their backbone architecture. 
One-stage approaches~\cite{ren2017accurate,liu2018learning,lin2018graininess,noh2018improving} where proposal generation and classification are formulated as a single-stage regression problem aim to accelerate the inference process of detectors, to meet the requirement of time efficiency in diverse real-world applications. In contrast to one-stage approaches, two-stage detectors aim to pursue the state-of-the-art performance by separate proposal generation followed by confidence computation of proposals. In recent years, two-stage pedestrian detection approaches~\cite{zhang2016faster,cai2016unified,mao2017can,song2018small,brazil2017illuminating,zhang2017citypersons,zhang2018occluded,brazil2019pedestrian} have shown superior performance on standard pedestrian benchmarks. For example, in~\cite{zhang2016faster}, RPN is employed to generate proposals and provide CNN features followed by a boosted decision forest. Zhang et al.~\cite{zhang2017citypersons} apply five key strategies to adapt the plain Faster R-CNN for pedestrian detection. Due to their superior performance on some pedestrian benchmarks~\cite{zhang2017citypersons}, we also deploy a two-stage detection method as a backbone pipeline in this work.

\subsection{Occlusion Handling in Pedestrian Detection}
Many efforts have been made to handle occlusions for pedestrian detection. A common strategy~\cite{brazil2019pedestrian,zhou2014non,tian2015deep,wang2018repulsion,ouyang2013joint,zhou2017multi} is the part-based approach where a set of part detectors are learned with each part designed to handle a specific occlusion pattern. The parts used in these approaches are usually manually designed, which may not be optimal. 

Different from the above approaches, there are also some other approaches~\cite{leibe2005pedestrian,wang2009hog,wang2018repulsion,zhang2018occlusion} for occlusion handling without using parts information. In~\cite{leibe2005pedestrian}, an implicit shape model is adopted to generate a set of pedestrian proposals which are further refined by exploiting local and global cues. Repulsion Loss~\cite{wang2018repulsion} and AggLoss~\cite{zhang2018occlusion} design two novel regression losses to generate more compact proposals to make them less sensitive to the NMS threshold. Besides, in~\cite{liu2019adaptive}, an adaptive NMS strategy is introduced that applies a dynamic suppression threshold to an instance in crowded scenes.

Contrary to the aforementioned methods, recent approaches focus on utilizing annotations of the visible body as extra supervisions together with the standard full body annotations to investigate the problem of occluded pedestrian detection. Zhang et al.~\cite{zhang2018occluded} employ visible body information along with a pre-trained body part prediction model to learn specific occlusion patterns (full, upper-body, left-body, and right-body visible). MGAN~\cite{pang2019mask}, a one-way supervision network, incorporates attention mechanisms into pedestrian detection using visible region supervision to emphasize the visible regions while suppressing the occluded regions. The work of Bi-box~\cite{zhou2018bi} regresses full and visible body of a pedestrian at the same time. However, the two branches of Bi-box~\cite{zhou2018bi} are trained separately with only score-level fusion, which cannot guarantee the detectors to learn robust enough pedestrian features.

In this work, we follow the idea of utilizing extra visible annotations to tackle the problem of occluded pedestrian detection. Different to~\cite{zhou2018bi}, our proposed method effectively integrates the two branches in the feature level to obtain more discriminative and robust features. Different to~\cite{pang2019mask}, our proposed method adopts a mutual-supervised way to make better use of the contextual features of occluding parts, aiming at enhancing the feature representations against heavy occlusions.

%% file: method.tex
\section{Proposed Method} 
In this paper, we propose a novel Mutual-Supervised Feature Modulation Network (MSFMN) for occluded pedestrian detection. The overall architecture of the proposed network is described in Sec.A. To obtain the most focused positive training samples, we propose a novel proposal sampling method in Sec.B. Next, we detail the design of the novel Mutual-Supervised Feature Modulation Module in Sec.C. Finally, the total loss function of multi-task prediction for end-to-end training along with a fusion method of two branches is represented during inference in Sec.D.

\subsection{Overall Architecture}
The overall architecture of the proposed method is illustrated in Fig. \ref{fig:overall}. Note that our proposed method can be easily applied to any existing two-stage detection frameworks. For a fair comparison, we implement it on the widely used Faster R-CNN framework~\cite{ren2015faster} and adopt VGG-16~\cite{simonyan2014very} as the backbone which is the most commonly used backbone in pedestrian detection networks. The architecture takes a raw image as input, first deploys a pre-trained ImageNet~\cite{deng2009imagenet} model. Then extracted feature maps are sent to the region proposal network (RPN) to generate two different sets of candidate proposals, including full body proposals and visible body proposals. For each proposal, a fixed-sized feature representation is obtained through RoI Align~\cite{he2017mask} layer. Finally, these features go through a classification network to generate predictions. Specifically, full body (FB) branch is a standard pedestrian detection branch to generate classification scores and regressed bounding box coordinates, and for the visible body (VB) branch, we need to consider the choice of employing classification task only or both classification and regression tasks. Ref.~\cite{cheng2018revisiting} discussed that classification needs translation invariant feature whereas regression needs translation covariant feature. If we design both classification and regression tasks in the VB branch, the regression task will force the detector to gradually learn translation covariant features during training, which might potentially downgrade the performance of the classifier. Therefore, we only place a classification task in the VB branch to accurately classify visible proposals. See Sec.IV-C for more comparative results.

\begin{figure}[!t]
\centering
\includegraphics[width=\linewidth]{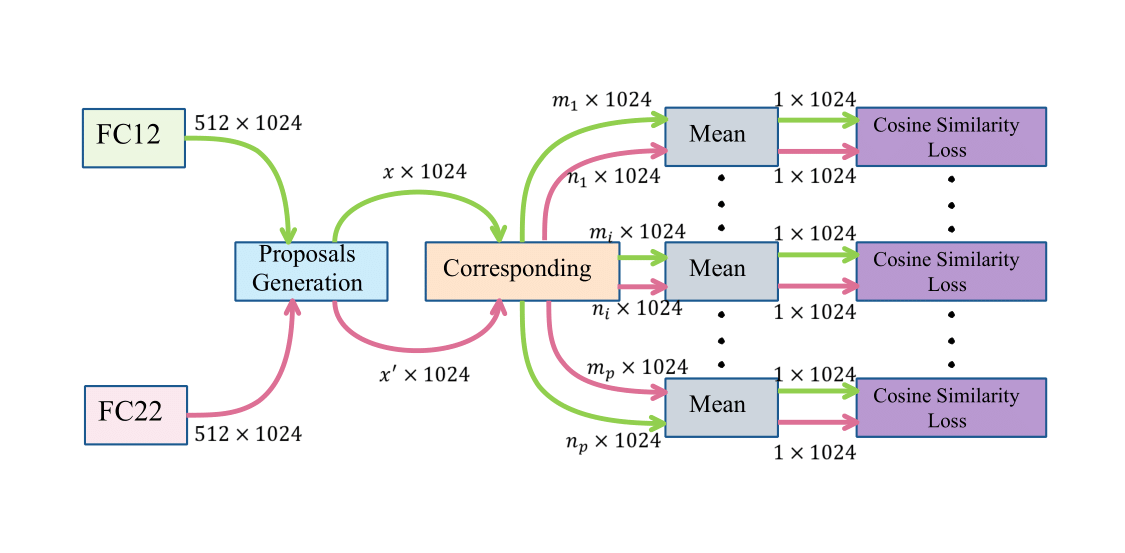}
\caption{The architecture of Mutual-Supervised Feature Modulation Module. It takes two 512 × 1024 feature vectors from FC12 and FC22, respectively. First, we collect $x$ and $x'$ positive training samples. Then, we find $m_i$ and $n_i$ training samples corresponding to the same pedestrian ground truth. P denotes the total number of pedestrians. Finally, we calculate cosine similarity loss of two 1 × 1024 feature vectors after $Mean$ operation.}
\label{fig:fig5_2}
\end{figure}

\subsection{Proposals Generation of RPN}
In our architecture, the FB branch collects training samples using full body annotations and then passes them through a classification network (FC11, FC12) to generate the classification scores and the regressed bounding box coordinates. VB branch collects training samples employing visible region annotations and then feeds them into a simple classification network (FC21, FC22) to obtain the classification scores, which indicate the probability that this visible proposal contains a pedestrian. The Proposals Generation process of RPN can be expressed with the following equations:
\begin{equation}\label{eq2}
\begin{split}
    P_{FB} =\{x|IoU(x,GT_{FB})>0.5\}   \cup \\ \{y|IoU(y,GT_{FB})<=0.5\}
\end{split}
\end{equation}

\begin{equation}
\begin{split}
P_{VB} =\{x’|IoU(x’,GT_{VB})>0.5\}   \cup  \\ \{y'|IoU(y',GT_{VB})<=0.5\}
\end{split}
\end{equation} where P$_{FB}$ and P$_{VB}$ represent the training samples collected for FB branch and VB branch respectively, $x$ and $x'$ denote positive training samples, and $y$ and $y'$ are negative training samples, respectively. For each branch, we sample 512 region proposals, the positive and negative samples are randomly sampled at a ratio of 1:3, following the same parameters as in~\cite{ren2015faster}.

\subsection{Mutual-Supervised Feature Modulation Module}
As shown in Fig. \ref{fig:fig5_2}, FC12 and FC22 are two fully connected layers of FB branch and VB branch, respectively. The 512 × 1024 feature vectors are passed to the Proposals Generation process stated in Sec.B to collect $x$ and $x'$ positive training samples. Then for two branches, we obtain $m_i$ and $n_i$ positive samples corresponding to the i-$th$ pedestrian ground truth. P represents the total number of pedestrians. To measure the distance between these $m_i$ and $n_i$ samples, we compare several different methods including Manhattan distance, Euclidean distance, and Cosine Similarity. Among these measurements, Cosine Similarity achieves the best results. For simplicity, we get two 1 × 1024 feature vectors by $Mean$ operation and then calculate the cosine similarity loss of these two vectors. See Sec.IV-C for more comparative results.

The cosine similarity loss is computed as:
\begin{equation}\label{eq}
L_{MSFMM} = \frac{1}{P}\sum_{i=1}^P[1- cos(\frac{1}{m_i}\sum v_i, \frac{1}{n_i}\sum v_i')]
\end{equation}where \textbf{$v_i$} and \textbf{$v_i'$} represent the $m_i$ × 1024 and  $n_i$ × 1024 feature vectors of the i-$th$ pedestrian extracted from FB and VB branch, respectively.

This mutual-supervised loss function is intended to enhance the feature representations by incorporating the visible part features and occluding part features simultaneously.

\subsection{Multi-task Optimization \& Inference}
Here, we present the loss function for the proposed architecture MSFMN. The overall loss formulation $L$ is as follows:
\begin{equation}
\begin{split}
L = L_{RPN_{cls}} + L_{RPN_{reg}} + L_{FB_{cls}} \\ + L_{FB_{reg}} + L_{VB_{cls}} +  L_{MSFMM}\label{eq}
\end{split}
\end{equation}
where $L_{RPN_{cls}}$ and  $L_{RPN_{reg}}$ refer to the classification and regression loss of RPN, $L_{FB_{cls}}$ and $L_{VB_{cls}}$ refer to the classification loss of FB and VB,  $L_{FB_{reg}}$ is the bounding box regression loss of FB and $L_{MSFMM}$ is the loss of Mutual-Supervised Feature Modulation Module.
Here, classification loss is Cross-Entropy loss and the bounding box regression loss is Smooth-L1 loss. 

In the inference stage, we propose a method to fuse the information of the two branches. Since the visible box contains the most discriminative information of pedestrians, taking the classification score of the VB branch as a part of the final score of detection box will improve the accuracy of pedestrian detector. Specifically, the classification scores of the VB branch are multiplied by those of the FB branch as the scores of the final detection box. Formally, the final scores of a pedestrian are defined as:
\begin{equation}
\begin{split}
Final\ Scores = Softmax(classification\ scores) \otimes \\
Softmax(visible\ classification\ scores)
\end{split}
\end{equation} 
where $classification\ scores$ and $visible\ classification\ scores$ represent the raw scores output from FC12 and FC22, respectively, $\otimes$ denotes element-wise product operation.

%% file: experiment.tex
\section{Experimental Evaluation}

\newcommand{\tabincell}[2]{\begin{tabular}{@{}#1@{}}#2\end{tabular}} 
In this section, we evaluate the proposed approach on CityPersons~\cite{zhang2017citypersons} and Caltech~\cite{dollar2011pedestrian}, which are two challenging pedestrian detection datasets with different occlusion settings.

\subsection{Datasets and Evaluation Metrics}
\textbf{Datasets:}
CityPersons~\cite{zhang2017citypersons} consists of 2975 training, 500 validation, and 1525 test images. It is a challenging dataset for pedestrian detection, which exhibits large diversity. 
Caltech pedestrian is a popular dataset~\cite{dollar2011pedestrian} containing 11 sets of videos. The first six video sets S0-S5 are for training and the remaining five video sets S6-S10 are used for testing. To increase the size of training set, we train the model on Caltech10×. Finally, the training and test sets have 42782 and 4024 images, respectively. 
Both datasets provide box annotations for full body and visible region.

\textbf{Evaluation Metrics:}
We report the performance using log-average miss rate (MR) throughout the experiments. It is computed over the false positive per image (FPPI) range of [10$^{-2}$, 10$^0$]~\cite{dollar2011pedestrian}, the lower value represents better detection performance. 
On Cityperons, we follow~\cite{zhang2017citypersons} and report the results across two different subsets: Reasonable (\textbf{R}), Heavy Occlusion (\textbf{HO}). For the Caltech dataset, we report results on Reasonable (\textbf{R}), Heavy Occlusion (\textbf{HO}), and the combined Reasonable + Heavy Occlusion (\textbf{R+HO}). The visibility ratio in \textbf{R} set is larger than 65\%, and the visibility ratio in \textbf{HO} set ranges from 20\% to 65\%. Thus, the visibility ratio in \textbf{R+HO} set is larger than 20\%. In all subsets, the height of pedestrians over 50 pixels is taken for evaluation, as in~\cite{zhang2018occluded}. Notice that \textbf{HO} set is designed to evaluate the performance under severe occlusions.

\begin{table}[!t]
\caption{Comparison(in log-average miss rates) of the MSFMN with the baseline on the CityPersons.}
\begin{center}
\begin{tabular}{|c|c|c|c|c|c|}
\hline
\textbf{\textit{Method}}& \textbf{\textit{VB Branch}}& \textbf{\textit{MSFMM}}& \textbf{\textit{R}} & \textbf{\textit{HO}} \\
\hline
Baseline   & $\times$   &   $\times$   &   11.92   &   47.88     \\
    \hline

\multirow{4}*{\textbf{Our MSFMN}} 

                        & cls+reg   &   $\times$   &11.45   &   45.35      \\
                        \cline{2-5}
                        &    cls  &   $\times$  &10.78  &  44.81  \\ \cline{2-5}
                       &    cls  & pos+neg &   10.52  &   41.35 \\ \cline{2-5}
                        &    cls  & pos &   \textbf{10.12}  &   \textbf{38.45}  \\
 \hline
    
\end{tabular}
\label{tab1}
\end{center}
\end{table}

\subsection{Implementation Details }
For both datasets, the network is trained on two GPUs with a total of 2 images per mini-batch. We adopt RoI Align~\cite{he2017mask} instead of RoI Pooling for feature extraction to get more precise features. We now detail settings specific to the two datasets.

\textbf{Citypersons}. We fine-tune pre-trained ImageNet VGG model~\cite{simonyan2014very} on the trainset of the CityPersons. We follow the same experimental protocol as in ~\cite{zhang2017citypersons} and employ two fully connected layers with 1024 instead of 4096 output dimensions. We choose SGD with momentum of 0.9 as the optimizer and set the initial learning rate as 0.0025. We train 15 epochs in total and decrease the learning rate by 0.1 at the 8-$th$ and 11-$th$ epochs.
As in~\cite{zhou2018bi}, ground-truth pedestrian examples which are at least 50 pixels tall and are occluded less than 70\% are used for training.

\textbf{Caltech}. We start with a model pre-trained on CityPersons dataset. The initial learning rate is 0.0025 for the first 3 epochs and is reduced by 10 and 100 times for another 2 and 1 epochs.

Multi-scale training and testing are not applied to ensure fair comparisons with previous methods. 


\subsection{Ablation Study}
To sufficiently verify the effectiveness of the proposed components, we conduct detailed ablation studies on CityPersons dataset.

\textbf{Baseline Comparison. } Tab. \ref{tab1} shows the performance of baseline and our proposed method on CityPersons validation subsets. For a fair comparison, we use the same training data, input scale (×1.3), and network backbone (VGG-16). The best results are boldfaced and shown in the final row. The baseline detector obtains a log-average miss rate of 11.92\% on \textbf{R} set of CityPersons dataset, outperforming the adapted FasterRCNN baseline in CityPersons~\cite{zhang2017citypersons} by 0.89\%. Thus, our baseline models are strong enough to verify the effectiveness of the proposed components. To analyze the contributions of the proposed components individually, we gradually apply the VB branch and Mutual-Supervised Feature Modulation Module (MSFMM) to the baseline model. As shown in Tab. \ref{tab1}, our final MSFMN significantly reduces the miss rates on both \textbf{R} and \textbf{HO} subsets. Under heavy occlusions (\textbf{HO}), MSFMN achieves an absolute reduction of 9.43\% in log-average miss rate compared to the baseline, demonstrates the effectiveness of MSFMN towards handling heavy occlusions.

The relevant ablation studies and analyses are presented in the following.

\textbf{Influence of VB branch. } To evaluate the efficacy of the proposed VB branch, we first add a VB branch based on the baseline model. We not only explore the effect of using classification task only in the VB branch but also explore the effect of using classification and regression tasks simultaneously. The results in Tab. \ref{tab1} validate our analyses in Sec.III-A. Specifically, the two branches network with only a classification task in VB branch outperforms baseline model on both \textbf{R} and \textbf{HO} sets by achieving a log-average miss rate of 10.78\% and 44.81\%, respectively.

\textbf{Influence of different proposals for Mutual-Supervised Feature Modulation Module.} We then apply Mutual-Supervised Feature Modulation Module for two-branch detector consisting of FB branch and VB branch (with classification task only) to demonstrate its effectiveness.
We not only consider applying similarity loss for positive samples but also consider applying similarity loss for both positive and negative samples. However, negative proposals have low similarity since they usually contain vastly different features. Therefore, it is more reasonable to calculate similarity loss for positive samples. Tab. \ref{tab1} confirms that calculating similarity loss of positive samples achieves a log-average miss rate of 10.12\% and 38.45\% on \textbf{R} and \textbf{HO}, respectively. 

\textbf{Influence of different similarity loss functions.} Next, we investigate the effect of different similarity loss functions, including Manhattan distance, Euclidean distance, and Cosine Similarity. The Manhattan distance is the distance between two points measured along axes at right angles, and Euclidean distance is the ``ordinary'' straight-line distance between two points in Euclidean space. And Cosine Similarity is to measure the difference between two individuals by cosine value of the angle between two vectors in vector space. As shown in Tab. \ref{tabloss}, Cosine Similarity method outperforms both Manhattan distance and Euclidean distance methods on both \textbf{R} and \textbf{HO} sets by achieving a log-average miss rate of 10.12\% and 38.45\%, respectively.


\begin{table}[!t]
\caption{Comparison (in log-average miss rates) of Different Similarity Method.}
\begin{center}
\begin{tabular}{|c|c|c|c|}
\hline
\textbf{\textit{Method}} & \textbf{\textit{R}} & \textbf{\textit{HO}} \\
\hline
Manhattan distance    & 11.25 &   46.79    \\ \hline
Euclidean distance    &  10.62 &   40.61   \\ \hline

Cosine Similarity  &   \textbf{10.12}   &   \textbf{38.45}   \\ \hline

\end{tabular}
\label{tabloss}
\end{center}
\end{table}

\begin{table}[!t]
\caption{Comparison (in log-average miss rates) With State-Of-The-Art on The CityPersons Val. Set.}
\begin{center}

\begin{tabular}{|c|c|c|c|}
\hline
\textbf{\textit{Method}}  & \textbf{\textit{Backbone}} & \textbf{\textit{R}} & \textbf{\textit{HO}}\\
\hline
Adaptive Faster RCNN~\cite{zhang2017citypersons} & VGG-16   &   12.81   &   - \\ \hline
Rep.Loss~\cite{wang2018repulsion} &ResNet-50  &   11.60   &   55.3 \\ \hline
Bi-box~\cite{zhou2018bi} & VGG-16 &   11.24   &   44.15 \\ \hline
FRCN+A+DT~\cite{zhou2019discriminative} & VGG-16   &   11.10   &   44.30 \\ \hline
OR-CNN~\cite{zhang2018occlusion}  & VGG-16  &   11.00   &   51.30\\ \hline
Adaptive NMS~\cite{liu2019adaptive} & VGG-16   &   10.80   &   54.00 \\ \hline
MGAN~\cite{pang2019mask} & VGG-16 &   10.50   &  39.40  \\ \hline
\textbf{Our MSFMN}  & VGG-16  &   \textbf{10.12}   &   \textbf{38.45}   \\ \hline
\end{tabular}
\label{tab6}
\end{center}
\end{table}

\subsection{State-of-the-art Comparison on CityPersons}
\begin{figure*}[t!]
\centering
\includegraphics[scale = 0.46]{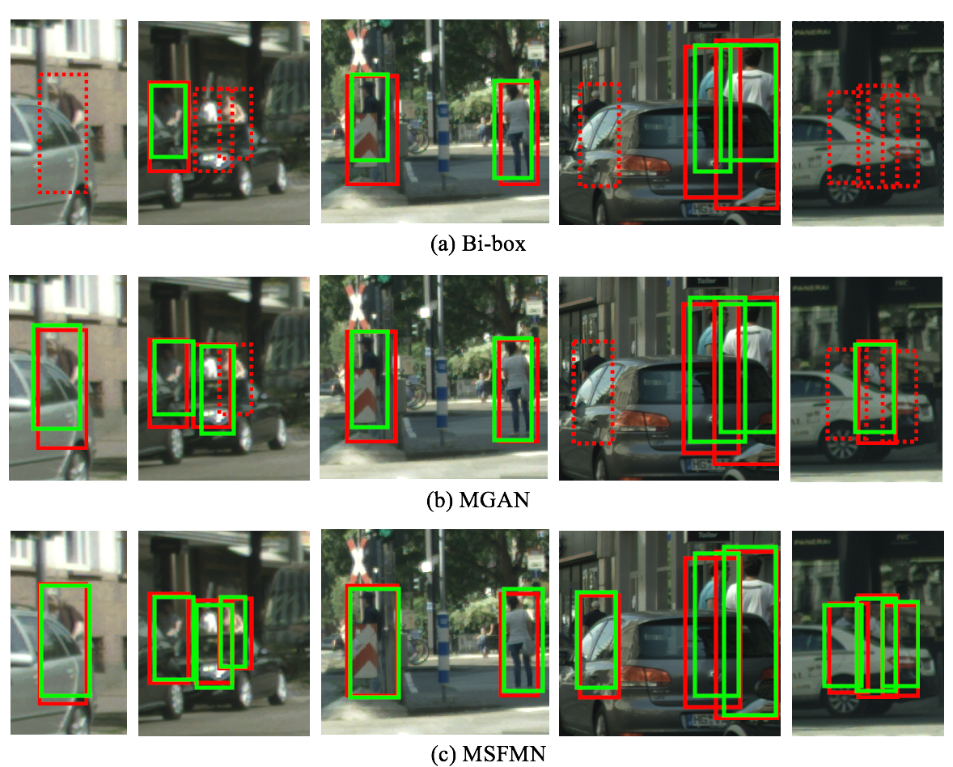}
\caption{Qualitative detection examples using (a) Bi-box~\cite{zhou2018bi}, (b) the state-of-the-art MGAN~\cite{pang2019mask} and (c) MSFMN on CityPersons val. images. Solid red boxes denote the ground-truth, dashed red boxes represent the missed detections and detector predictions are indicated by green boxes. The detected regions are cropped from the corresponding images for improved visualization. Note that all detection results are obtained using the same false positive per image (FPPI) criterion. Our MSFMN accurately detects pedestrians with varying levels of occlusions.}
\label{fig:fig6}
\end{figure*}

We compare our method with other recent state-of-the-art methods including Adapted FasterRCNN~\cite{zhang2017citypersons}, Rep. Loss~\cite{wang2018repulsion}, OR-CNN~\cite{zhang2018occlusion}, Bi-box~\cite{zhou2018bi}, Adaptive NMS~\cite{liu2019adaptive}, FRCN+A+DT~\cite{zhou2019discriminative} and MGAN~\cite{pang2019mask} on CityPersons dataset. As shown in Tab. \ref{tab6} , we report the performance of MSFMN and other methods on the validation set using the same ground-truth pedestrian examples and input scale during training. The proposed MSFMN outperforms all the other methods on \textbf{R} and \textbf{HO} subsets. Notably, on \textbf{HO}, our method reduces the MR of state-of-the-art result from 39.40\% to 38.45\%, demonstrating the superiority of the proposed method in heavy occlusion cases. Fig. \ref{fig:fig6} displays example detections from Bi-box~\cite{zhou2018bi}, MGAN~\cite{pang2019mask}, and the proposed MSFMN on CityPersons val. set. The occlusion degrees of examples vary widely from partial to heavy occlusion. Our MSFMN accurately detects pedestrians with varying levels of occlusions.


\subsection{State-of-the-art Comparison on Caltech }
\begin{figure*}[htp]
\centering
\includegraphics[scale = 0.46]{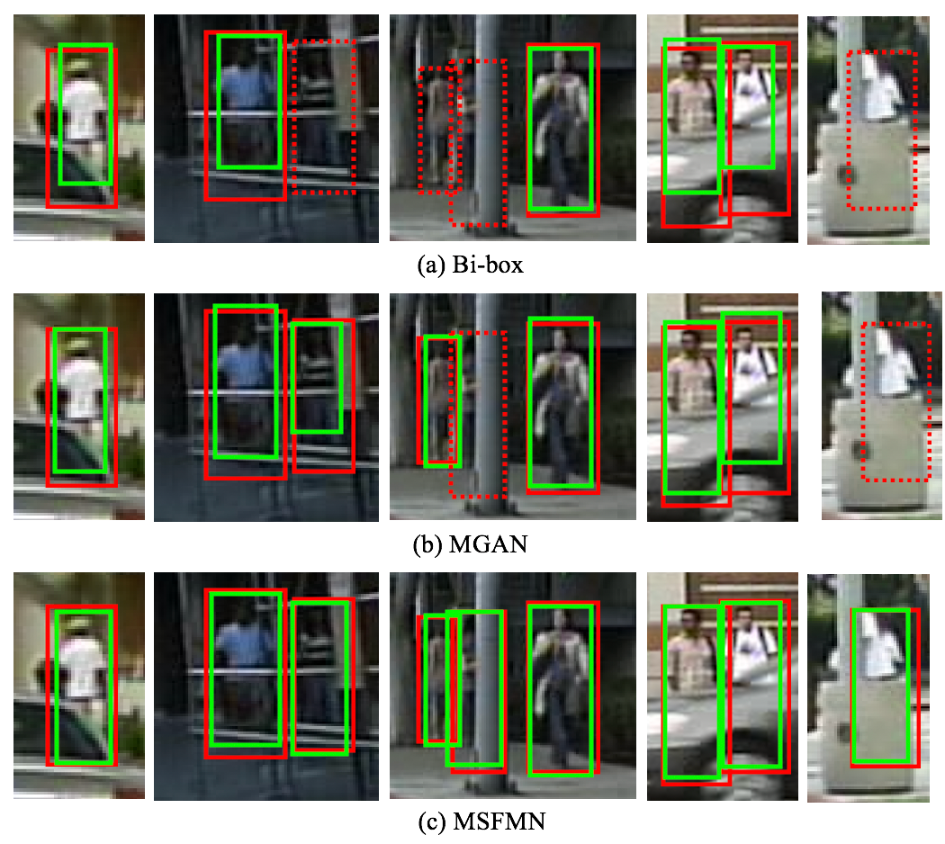}
\caption{Qualitative detection comparison of (a) Bi-box~\cite{zhou2018bi}, (b) MGAN~\cite{pang2019mask} and (c) MSFMN under different occlusions on caltech test images. All detection results are obtained using the same false positive per image criterion. The solid red boxes denote the ground-truth, dashed red boxes represent the missed detections and the green boxes present detection results. For better visualization, the detected regions are cropped from the corresponding images.}
\label{fig:figcaltech}
\end{figure*}

Finally, we evaluate the MSFMN on Caltech~\cite{dollar2011pedestrian} and compare it with state-of-the-art approaches. Tab. \ref{tab8} shows the comparison on Caltech test set under three occlusion subsets: \textbf{R}, \textbf{HO}, and \textbf{R + HO}. $*^O$ means the result is under the standard (old) test annotations, and $*^N$ means the result is under the new annotations provided by~\cite{zhang2016how}. 
Compared to existing methods, the MSFMN achieves superior detection performance on all these subsets with log-average miss rate of 6.45\%, 38.01\%, 13.40\%, and 2.80\%, respectively. Fig. \ref{fig:figcaltech} depicts the detection examples of the proposed MSFMN and Bi-box~\cite{zhou2018bi} and MGAN~\cite{pang2019mask}. Our proposed MSFMN provides more accurate detections under different occlusion scenarios compared to the other two approaches.

\begin{table}[!t]
\caption{Comparison (in log-average miss rates) with State-Of-The-Art Comparison on Caltech Test Set.}
\begin{center}
\begin{tabular}{|c|c|c|c|c|}
\hline
\textbf{\textit{Method}} & \textbf{\textit{R$^O$}} & \textbf{\textit{HO$^O$}} & \textbf{\textit{R+HO$^O$}} & \textbf{\textit{R$^N$}}\\
\hline

DeepParts~\cite{tian2015deep}   &   11.89 & 60.42   &   22.79   &12.90   \\  \hline
MS-CNN~\cite{cai2016unified}    &   9.95  &  59.94   &   21.53 &   8.08  \\  \hline
ATT-part~\cite{zhang2018occluded}     &   10.33  &   45.18   &   18.21 &   8.11 \\  \hline
SDS-RCNN~\cite{brazil2017illuminating}   &   7.36    &   58.55   &19.72    &   6.44  \\  \hline
OR-CNN~\cite{zhang2018occlusion}    & - &-&-&  4.10   \\  \hline
Rep.Loss~\cite{wang2018repulsion}  & - &-&-&  4.00  \\  \hline
Bi-box~\cite{zhou2018bi}  &  7.61 &44.40   &   16.06   & -       \\ \hline
MGAN~\cite{pang2019mask}  &   6.83 & 38.16 &   13.84   &   -   \\ \hline
\textbf{Our MSFMN}  &   \textbf{6.45} &  \textbf{38.01} & \textbf{13.40}  & \textbf{2.80} \\ \hline

\end{tabular}
\label{tab8}
\end{center}
\end{table}

%% file: conclusion.tex
\section{Conclusion}
We propose a novel Mutual-Supervised Feature Modulation Network (MSFMN) for occluded pedestrian detection. A new Mutual-Supervised Feature Modulation Module is designed to calculate the similarity loss of full body boxes and visible body boxes corresponding to the same pedestrian aiming at enhancing feature representations of heavily occluded pedestrians. Our MSFMN consists of two branches, a standard full body detection branch and an extra visible body classification branch. Moreover, we utilize the full body annotations and visible annotations to supervise the two branches respectively. The effectiveness of the proposed MSFMN is validated on the CityPersons and Caltech datasets. Experimental results demonstrate that the proposed MSFMN outperforms other state-of-the-art approaches, validating the effectiveness of the proposed method.

\textbf{Future work}: the final score generation in the inference stage is relatively simple, therefore we plan to explore more effective ways to combine the information of two branches in the future.

%% file: acknowledgment.tex
\section*{Acknowledgment}
This work was partly supported by National Natural Science Foundation of China under Grants 61703039 and 62072032, and Beijing Natural Science Foundation under Grant 4174095.